\documentclass{article}

\usepackage{arxiv}

\usepackage[utf8]{inputenc} % allow utf-8 input
\usepackage[T1]{fontenc}    % use 8-bit T1 fonts
\usepackage{hyperref}       % hyperlinks
\usepackage{url}            % simple URL typesetting
\usepackage{booktabs}       % professional-quality tables
\usepackage{amsfonts}       % blackboard math symbols
\usepackage{nicefrac}       % compact symbols for 1/2, etc.
\usepackage{microtype}      % microtypography
\usepackage{lipsum}		% Can be removed after putting your text content
\usepackage{graphicx}
\usepackage{doi}
\usepackage{natbib}

\usepackage{subfigure}
\usepackage{amsmath}
\usepackage{amssymb}
\usepackage{booktabs}
\usepackage{colortbl}
\usepackage{multirow}
\usepackage{threeparttable}
\usepackage{pifont}
\usepackage{makecell}
\newcommand{\etal}{\textit{et al.}}

\definecolor{babyblueeyes}{rgb}{0.63, 0.79, 0.95}
\definecolor{babyblue}{rgb}{0.54, 0.81, 0.94}
\definecolor{columbiablue}{rgb}{0.61, 0.87, 1.0}
\newcommand{\cmark}{\ding{51}}%
\newcommand{\xmark}{\ding{55}}%

\title{TEAM-Net: Multi-modal Learning for Video Action Recognition with Partial Decoding}

\date{} 					% Or removing it

\author{ Zhengwei Wang\thanks{Accepted by BMVC 2021} \\
    V-SENSE\\
    School of Computer Science and Statistics\\
    Trinity College Dublin\\
    Ireland\\
	\texttt{villa.wang.zhengwei@gmail.com} \\
	\And
	Qi She \\
    ByteDance\\
    China\\
	\texttt{sheqi1991@gmail.com} \\
	\And
	Aljosa Smolic \\
    V-SENSE\\
    School of Computer Science and Statistics\\
    Trinity College Dublin\\
    Ireland\\
	\texttt{smolica@tcd.ie} \\
}

\begin{document}

\maketitle

\begin{abstract}
Most of existing video action recognition models ingest raw RGB frames. However, the raw video stream requires enormous storage and contains significant temporal redundancy. Video compression (e.g., H.264, MPEG-4) reduces superfluous information by representing the raw video stream using the concept of Group of Pictures (GOP). Each GOP is composed of the first I-frame (aka RGB image) followed by a number of P-frames, represented by motion vectors and residuals, which can be regarded and used as pre-extracted features. In this work, we 1) introduce sampling the input for the network from partially decoded videos based on the GOP-level, and 2) propose a plug-and-play mul\textbf{T}i-modal l\textbf{EA}rning \textbf{M}odule (TEAM) for training the network using information from I-frames and P-frames in an end-to-end manner. We demonstrate the superior performance of TEAM-Net compared to the baseline using RGB only. TEAM-Net also achieves the state-of-the-art performance in the area of video action recognition with partial decoding. Code is provided at \url{https://github.com/villawang/TEAM-Net}.
\end{abstract}

% keywords can be removed

%-------------------------------------------------------------------------
\section{Introduction}
\label{sec:intro}
Video understanding has drawn an increasing amount of attention, since video data accounts for 82\% of all Internet traffic by 2022~\citep{cisco2018cisco}. For instance, millions of videos are uploaded to TikTok, Douyin, and Xigua Video to be processed everyday, wherein understanding video content acts a pivotal part. One of the most important tasks in video understanding is to understand human actions, which has many real-world applications, such as Virtual Reality/Augmented Reality (VR/AR) and human computer interaction. Video data contains a rich source of visual content including appearance information in each individual frame and motion information across consecutive frames, which poses challenges for effectively extracting information from video data.

Traditional action recognition approaches~\citep{carreira2017quo,feichtenhofer2019slowfast,lin2019tsm,li2020tea,Wang_2021_CVPR} investigated better modeling video data either by designing new architectures~\citep{feichtenhofer2019slowfast} or embedded modules~\citep{lin2019tsm}. These architectures and embedded modules are designed for highlighting the spatio-temporal perspective contained in videos. Most of these approaches are directly designed for images to parse a video frame by frame. However, video data contains significant redundancy both on the spatial and the temporal dimension. For instance, an 1h of 720p video can be compressed from 222GB raw to 1GB~\citep{wu2018compressed}, which indicates redundant information contained in videos. 
% Temporal Segment Network (TSN)~\citep{wang2016temporal} introduced a well-know sparse sampling strategies for selecting frames from videos to train the network, which extracted short snippets over a long video sequence with a uniform sparse sampling, but it still did not stress the redundancy of frames in a video sequence. Recent approaches~\citep{zhu2016key,kar2017adascan,dong2019attention,zheng2020dynamic} investigated key frames selection in video data. However, these approaches require an extra model to learn useful frames, which is computationally expensive especially in the inference phase. 

\begin{figure}[!htpb]
    \centering
    \includegraphics[width=1.\textwidth]{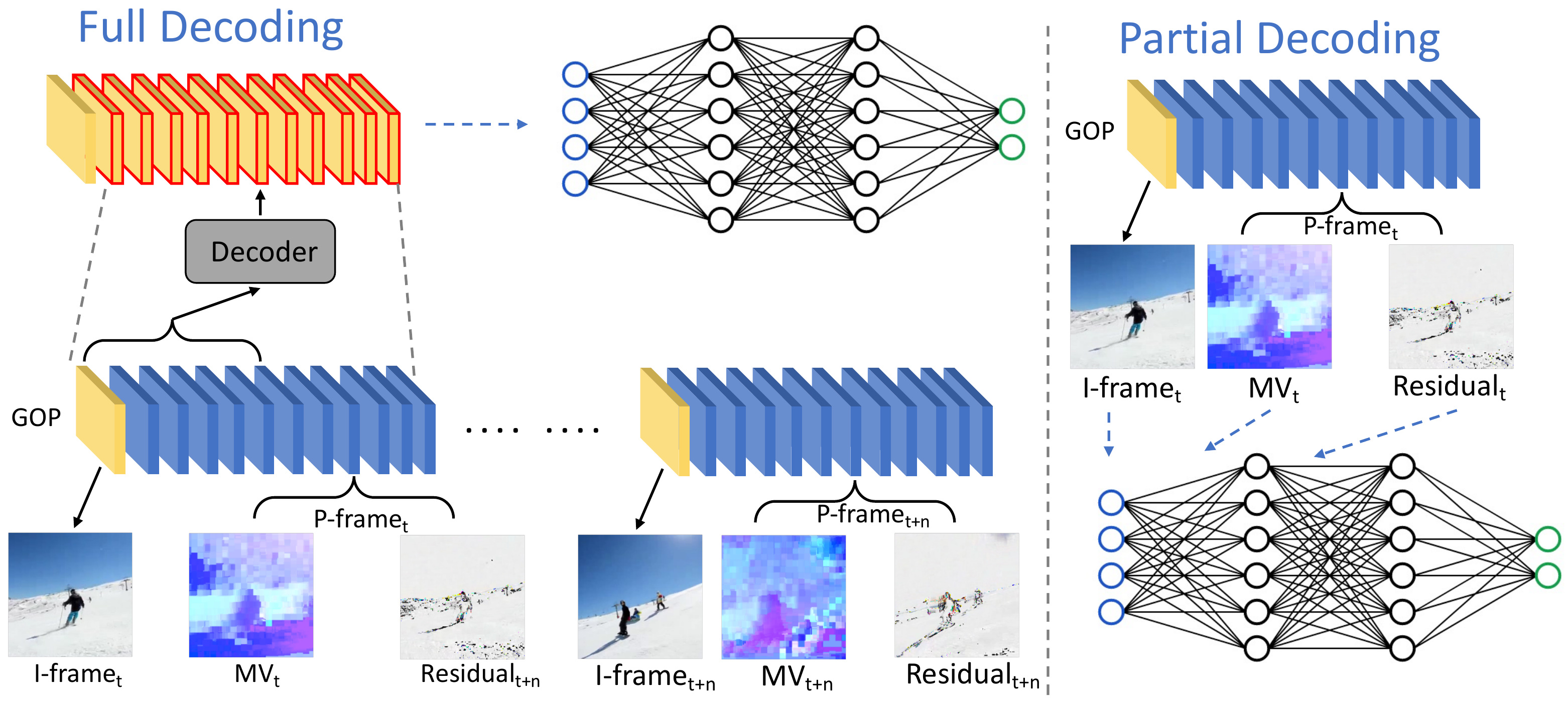}
    \vspace{-10pt}
    \caption{Comparison between traditional video action recognition (left) and partially decoded video action recognition (right). Left: Decoded RGB images are fed to the network. RGB images highlighted in red are reconstructed from the first I-frame and consecutive P-frames~\citep{wu2018compressed} in the corresponding GOP. Right: Without decoding every RGB image in a GOP, the partial decoding uses the first decoded I-frame and partial information from one P-frame (including MV and residual) in a GOP as an input.  
    }
    \vspace{-8pt}
    \label{fig:GOP}
\end{figure}

Performing action recognition in the \textit{partially compressed domain} becomes a recent interest~\citep{wu2018compressed,shou2019dmc,li2020slow,huo2020lightweight,battash2020mimic}. The difference between performing in the raw domain (full decoding) and partially compressed domain (partial decoding) is illustrated in Figure~\ref{fig:GOP}. In terms of partial decoding on the right, video data is parsed by a stream of Group of Pictures (GOPs). Each GOP starts with an intra-frame (I-frame, aka an independently encoded RGB image) followed by $n$ P-frames ($n = 11$ in this work). Regarding full decoding, RGB images (highlighted in red on the left top) can be reconstructed from the first I-frame and following consecutive P-frames in the corresponding GOP shown on the bottom left~\citep{wu2018compressed}. In other words, highlighted RGB images in red \textit{only} depend on the first I-frame and a number of P-frames in the current GOP. Traditional action recognition approaches randomly sample from these RGB images. However, it would make more sense to sample images from different GOPs and to perform on the GOP-level. Further, on the GOP-level, we are able to take benefits from additional information provided by P-frames i.e., two extra modalities Motion Vectors (MVs) and residuals. These two modalities represent motion and texture/color change information between the current decoded RGB image and its reference I-frame, which are very useful for action reasoning. This information can be regarded and used as pre-extracted features to improve the network performance. It should be noticed that previous works~\citep{wu2018compressed,shou2019dmc,li2020slow,huo2020lightweight,battash2020mimic} term this type of action recognition as \textit{`compressed video action recognition'}. We argue the terminology \textit{`compressed'} is not fully correct because original compressed video formats such as MPEG-4 and H.264 stored in the Video Coding Layer (VCL) do not contain decoded frames i.e., they are compressed bit streams~\citep{wiegand2003overview}. All previous works used decoded I-frames, and computed representative images from decoded MVs and residuals for training the network but not the compressed bit stream data directly. Unlike traditional action recognition that decodes every RGB image, this operation decodes the first I-frame and some information from P-frames in each GOP. So we name such a domain for processing video data as \textit{`partially compressed domain'}, which refers to \textit{`compressed domain'} in previous works.

\begin{figure}[!htpb]
    \centering
    \includegraphics[width=1.\textwidth]{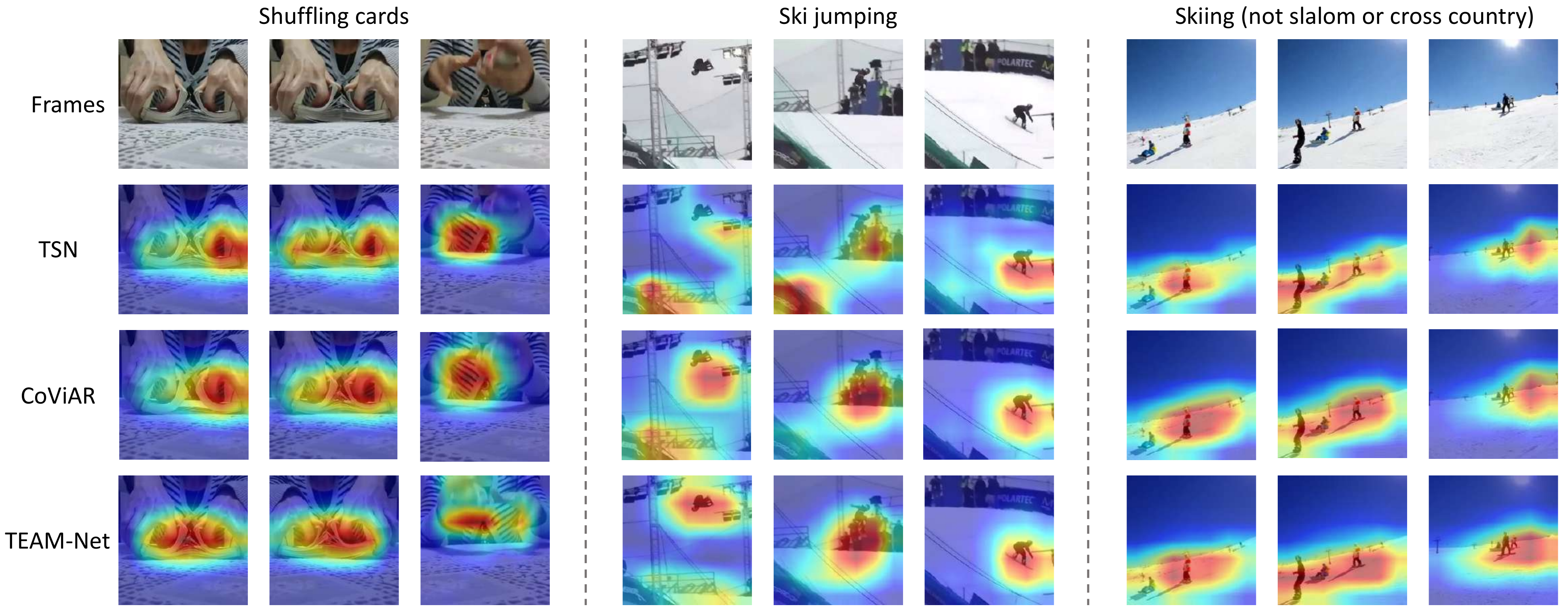}
    \vspace{-20pt}
    \caption{Visualization for significant features extracted by TSN, CoViAR and TEAM-Net. Features extracted by each method are visualized by using CAM~\citep{zhou2016learning}. It should be noted that our TEAM-Net does not contain any temporal modeling module for capturing motion information in videos, but uses the pre-extracted motion features from the MVs.}
    \label{fig:visualization}
\end{figure}

Previous works~\citep{wu2018compressed,shou2019dmc,li2020slow,huo2020lightweight,battash2020mimic} investigated adding two extra modalities MVs and residuals in the partially compressed domain to improve the model performance. Most of current approaches~\citep{wu2018compressed,shou2019dmc,huo2020lightweight} require training separated backbones for I-frames, MVs and residuals respectively, in which fusion only comes at the inference stage. This operation causes expensive computation for training models especially for a large-scale dataset. The fusion capability is also constrained by omitting mid-level fusion during training. In this work, we propose a plug-and-play mul\textbf{T}i-modal l\textbf{EA}rning \textbf{M}odule (TEAM) to help train the network with three modalities by ingesting partially decoded videos in an end-to-end manner. We study the TEAM by adopting the widely adopted Temporal Segment Network (TSN) framework~\citep{wang2016temporal} in the partially compressed domain. It is worth nothing that our TEAM can be embedded to other advanced action recognition frameworks such as TSM~\citep{lin2019tsm}. To this end, we find the network equipped with our TEAM (TEAM-Net) outperforms the TSN in the raw domain i.e., using RGB only, and CoViAR in the partially compressed domain. Figure~\ref{fig:visualization} shows the visualization for well-trained TSN, CoViAR and TEAM-Net. It can be seen that TEAM-Net successfully makes use of MV and residual modalities for better action reasoning compared to the other two baselines. In summary, our contributions are three-fold:
\begin{itemize}
    \item We propose a novel plug-and-play module for fusing three modalities, which enables training the network with I-frames, MVs and residuals in an end-to-end manner for video action recognition with partial decoding.
    \item Unlike traditional action recognition that samples inputs on the frame-level, we introduce sampling inputs on the GOP-level for training the network, which can be understood as an extension of TSN in the partially compressed domain.
    \item Extensive experiments have been conducted to show that our proposed TEAM-Net outperforms the traditional action recognition baseline TSN and is superior to current state-of-the-art in video action recognition with partial decoding.
\end{itemize}

%-------------------------------------------------------------------------
\section{Related Works}
\textbf{Traditional Video Action Recognition.}\hspace{3pt} Recent video action recognition approaches normally operate in the raw video domain i.e., RGB images decoded from videos are fed to the network~\citep{wang2016temporal,tran2015learning,feichtenhofer2019slowfast}. Current approaches can be mainly divided into two categories 1) 3D CNN-based approaches (clip level)~\citep{tran2015learning,hara2018can,feichtenhofer2019slowfast}, and 2) 2D CNN with efficient temporal module design (frame level)~\citep{lin2019tsm,li2020tea,jiang2019stm,Wang_2021_CVPR}. Different from image data, videos contain both spatial information in each individual frame and temporal information in the sequence i.e., movements. 3D CNNs utilize 3D convolutions to characterize spatio-temporal information for the clip input directly (a clip contains a number of frames). I3D~\citep{carreira2017quo} inflated the ImageNet pre-trained 2D convolution to a 3D convolution. SlowFast networks~\citep{feichtenhofer2019slowfast} were proposed to model inconsistent action tempos e.g., \textit{`running vs walking'}. It comprised a slow path for modeling slow actions and a fast path for modeling fast actions respectively. While 3D CNN-based approaches have achieved exciting performance on several benchmark datasets, these models contain massive parameters and require expensive computation. Various problems may arise, such as overfitting~\citep{hara2018can}, difficulty in converging~\citep{tran2018closer} and slow inference~\citep{zhu2020comprehensive}. Even though recent works~\citep{qiu2017learning,tran2018closer} have demonstrated that the 3D convolution can be factorized to lessen computations to some extent, the
computation of 3D CNN-based models is still much more of a burden compared to 2D CNN-based models. Different from optimizing 3D CNNs on clips directly, 2D CNNs deploy 2D convolutions in a frame-level manner. TSN~\citep{wang2016temporal} applied 2D CNNs to video action recognition and introduced the concept of `\textit{segment}' to sample frames from videos i.e., extract short snippets over a long video sequence with a uniform sparse sampling scheme. Direct use of 2D CNNs is difficult to fully capture the temporal information of video. Current mainstream methods focus on designing an efficient module that can be embedded into 2D CNNs for temporal reasoning. TSM~\citep{lin2019tsm} introduced a shift operation for a part of channels along the temporal dimension. TEA~\citep{li2020tea} proposed a motion excitation module for short range temporal modeling and a multiple temporal aggregation module for long range temporal modeling. These two modules are connected sequentially. ACTION-Net~\citep{Wang_2021_CVPR} proposed a multipath excitation module for spatio-temporal, channel and motion modeling. These proposed modules can be inserted into 2D CNNs, in which temporal modeling is introduced to the network. 

\noindent
\textbf{Action Recognition from Partially Decoded Video.}\hspace{3pt} Apart from manually designing methods for extracting inherent information in videos, video data itself contains useful information that previous approaches aim to model, such as motion information. Two-stream~\citep{carreira2017quo} approaches have demonstrated that augmenting input data by adding optical flow~\citep{ilg2017flownet} is able to significantly enhance the performance for video action recognition. However, computation of optical flow is very expensive, which is not feasible for those applications requiring low latency in real life. On the other hand, exploiting information computed for encoding of video~\citep{smolic2000low} recently comes into focus for video action recognition~\citep{zhang2016real,zhang2018real,wu2018compressed,huo2020lightweight,li2020slow,shou2019dmc}. Previous works~\citep{wu2018compressed,huo2020lightweight,li2020slow,shou2019dmc} tackled video action recognition using I-frames, MVs and residuals. Works~\citep{zhang2016real,zhang2018real} extracted MVs from videos to replace the optical flow. Wu~\etal~\citep{wu2018compressed} proposed the CoViAR model that contained three 2D CNNs for ingesting three modalities respectively. It is worth nothing that these three 2D CNNs were trained independently, in which fusion of three modalities was only performed at the inference phase. Huo~\etal~\citep{huo2020lightweight} proposed an Aligned Temporal Trilinear Pooling (ATTP) module for better fusing three modalities at the inference stage. Due to the low resolution of MVs, the DMC-Net~\citep{shou2019dmc} applied Generative Adversarial Networks (GANs)~\citep{goodfellow2014generative,wang2021generative} for generating motion cues by adopting the information of MVs, residuals and optical flow. Inspired by SlowFast networks~\citep{feichtenhofer2019slowfast}, Li~\etal~\citep{li2020slow} proposed an end-to-end Slow-I-Fast-P (SIFP) framework that treated I-frames as the slow pathway and P-frames as the fast pathway. Battash~\etal~\citep{battash2020mimic} proposed an end-to-end framework that ingests a clip input as one GOP, in which a teacher-student distillation strategy was utilized for improving the model performance. IMRNet~\citep{IMRNet} employed bidirectional dynamic connections between I-frame \& MV pathways and between I-frame \& residual pathways. 

\noindent
\textbf{Multi-modal Action Recognition.}\hspace{3pt}
Previous works have studied learning of multiple streams present in video~\citep{kazakos2019epic,miech2018learning,nagrani2018seeing,liu2019use}. Most of these works deployed the audio stream as complementary information to enhance the performance. Our work makes use of inherent visual information, which exists in partially decoded video, and can be integrated to frameworks mentioned above. In this work, we tackle the multi-modal learning side that improves the network performance on action recognition by adopting multi-modal data in partially decoded videos. Previous works~\citep{li2020slow,battash2020mimic} investigated the feasibility of training models in the partially compressed domain in an end-to-end manner. \citep{li2020slow} is only suitable for a SlowFast-like architecture while the upper bound performance of \citep{battash2020mimic} the teacher model trained in the raw domain. Luvizon~\etal~\citep{luvizon20182d} proposed a multi-task framework jointly trained by pose estimation and action recognition tasks. PoseMap~\citep{liu2018recognizing} employed a two-stream network to process spatio-temporal pose heatmaps and inaccurate poses separately. Wang~\etal~\citep{wang2020catnet} presented a two-stream network for processing RGB and depth modalities in videos. A bilinear pooling block was investigated for RGB-D video action recognition in~\citep{hu2018deep}. Wang~\etal~\citep{wang2020makes} proposed gradient blending for avoiding overfitting in multi-modal networks.

\noindent
\textbf{Attention Mechanisms.}\hspace{3pt} Attention mechanisms have been widely adopted in computer vision tasks. Hu~\etal~\citep{hu2018squeeze} proposed a SE block for explicitly modeling channel interdependencies using the attention mechanism for the channel dimension. CBAM~\citep{woo2018cbam} was proposed beyond the SE block by adding an extra attention map along the spatial dimension. Joze~\etal~\citep{joze2020mmtm} investigated attention mechanisms for fusing multi-modal features from the channel perspective and proposed a MMTM module beyond the SE. Inspired by previous works~\citep{hu2018squeeze,woo2018cbam,joze2020mmtm}, we propose a mul\textbf{T}i-modal l\textbf{EA}rning \textbf{M}odule (TEAM) for fusing I-frames, MVs and residuals with respect to the channel dimension and the spatial dimension at different levels inside an end-to-end model. 
%------------------------------------------------------------------------
\section{Design of TEAM}
We first clarify notations to be used in this section: N (batch size), T (temporal length), C (channel), H (height), W (width) and r (reduction ratio). Given input feature maps to TEAM: $\mathbf{F}_1 \in \mathbb{R}^{NT\times C_1\times H\times W}$, $\mathbf{F}_2 \in \mathbb{R}^{NT\times C_2\times H\times W}$ and $\mathbf{F}_3 \in \mathbb{R}^{NT\times C_3\times H\times W}$, for three modalities I-frames, MVs and residuals respectively (the spatial size for three modalities is equal but the channel number is different due to different backbones that are used for different modalities), we aim to optimize the entire model jointly using three modalities. To achieve this, we construct joint feature representations along the channel and the spatial dimension to be processed by a channel fusion module and a spatial fusion module respectively. 
\begin{figure}[!htbp]
    \centering
    \subfigure[]{
        \includegraphics[width=.45\textwidth]{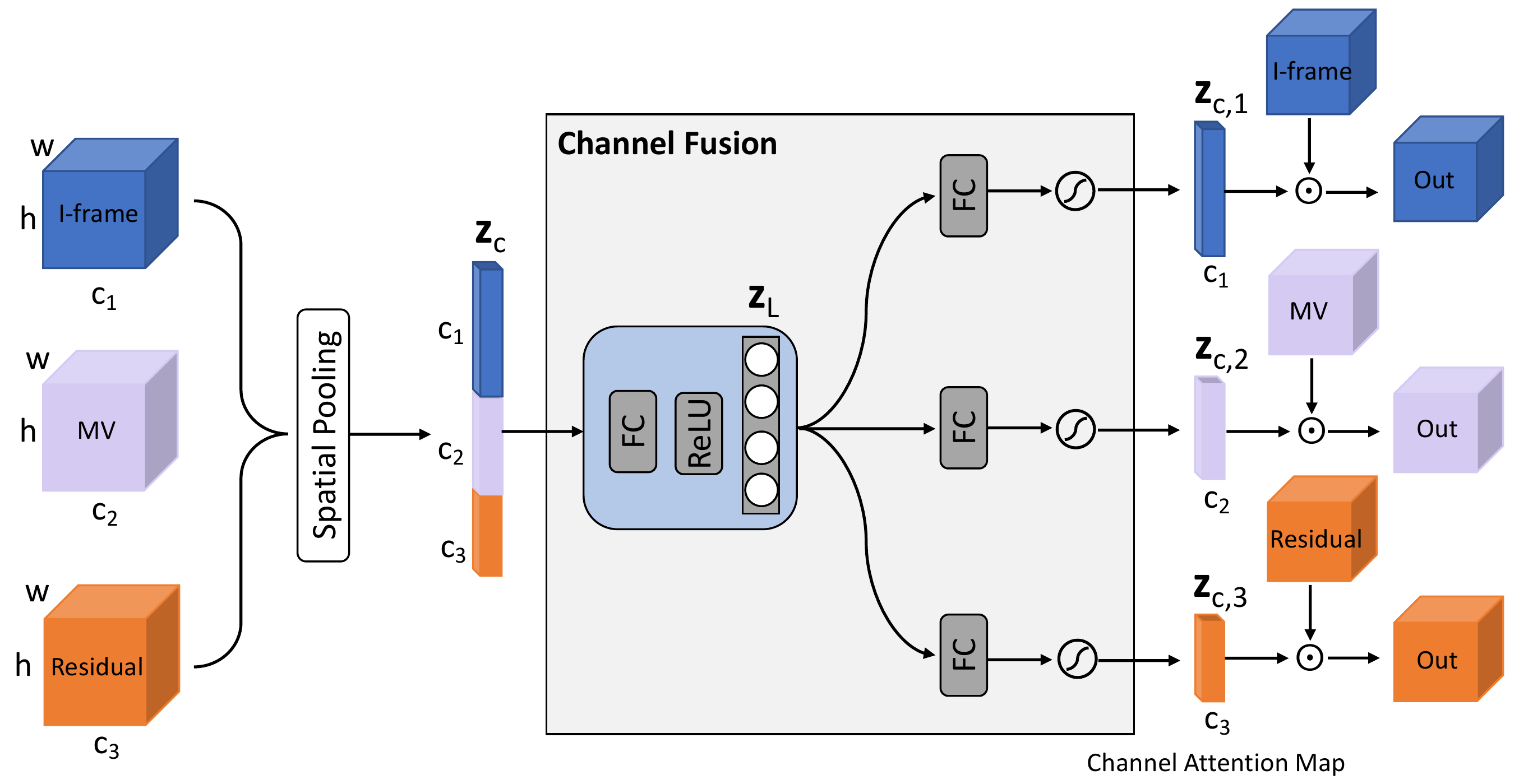}
        \label{fig:team_CF}    
    }
    \hfill
    \subfigure[]{
        \includegraphics[width=.5\textwidth]{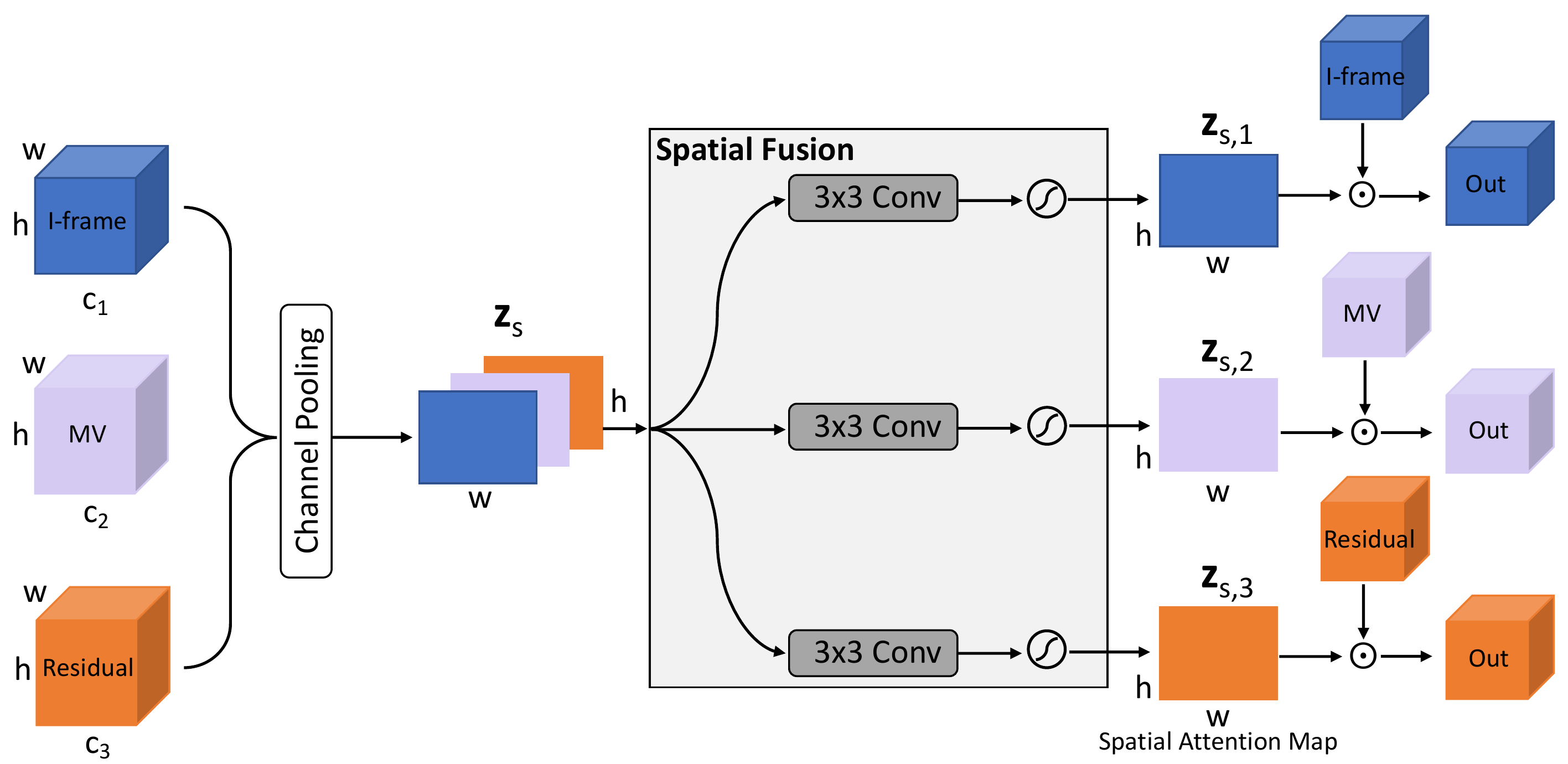}
        \label{fig:team_SF}    
    } 
    \subfigure[]{
        \includegraphics[width=.6\textwidth]{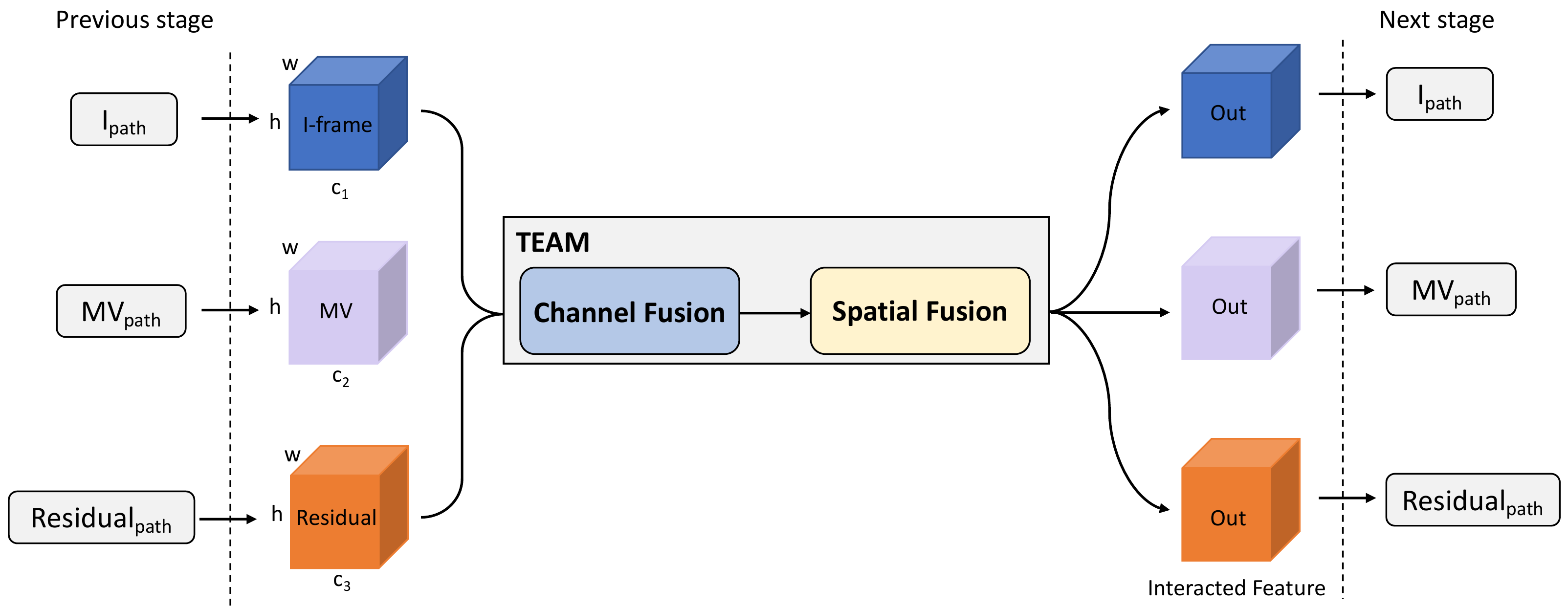}
        \label{fig:team_module}    
    } 
    \caption{Schematics of channel fusion, spatial fusion and TEAM module. The TEAM module in (c) involves two-stage fusion of (a) the channel fusion module and (b) the spatial fusion module.}
    \label{fig:team_diag}
\end{figure}

\noindent
\textbf{Channel Fusion.}\hspace{3pt} Figure~\ref{fig:team_CF} illustrates the channel fusion module. As we only care about channel information at this stage, we average each modality feature on the spatial dimension
\begin{equation}
\mathbf{F}_{c,i} = \frac{1}{H\times W}\sum^{H}_{m=1}\sum^{W}_{n=1}\mathbf{F}_{i}[:,:,m,n], 
    \label{eq:spatial_pool}
\end{equation}
where $\{c,i\}$ indicates the channel fusion (to be distinguished from the spatial fusion) and modality type i.e., $i=1$ for I-frames, $i=2$ for MVs, and $i=3$ for residuals, and $\mathbf{F}_i$ is the input for the corresponding modality. We then construct a joint channel representation $\mathbf{z}_c\in \mathbb{R}^{NT\times C_z}$ ($C_z = C_1+C_2+C_3$) by concatenating three feature maps along the channel dimension. We follow the same squeeze strategy as in~\citep{hu2018squeeze} to reduce the channel dimension for the joint channel representation as 
\begin{equation}
    \mathbf{z}_l = \text{ReLU}(\mathbf{z}_c \mathbf{W}_c),
    \label{eq:CF_squeeze}
\end{equation}
where $\mathbf{z}_l\in \mathbb{R}^{NT\times C_z/r}$ and $\mathbf{W}_c \in \mathbb{R}^{C_z\times C_z/r}$ is the FC weight. The squeezed representation $\mathbf{z}_l$ is then expanded by three FC layers followed by a Sigmoid to generate three attention maps for each modality as follows
\begin{equation}
    \mathbf{z}_{c,i} = \sigma(\mathbf{z}_l \mathbf{W}_{c,i}), 
\label{eq:CF_expand}
\end{equation}
where $\mathbf{z}_{c,i}\in \mathbb{R}^{NT\times C_i}$ is the attention map and $\mathbf{W}_{c,i} \in \mathbb{R}^{C_z/r\times C_i}$ is the FC weight for the corresponding modality.
$\sigma$ is the Sigmoid function. The final output for each modality can be computed as 
\begin{equation}
\begin{aligned}
    \mathbf{F}^{out}_{c,i} = \mathbf{F}_{i} \odot \mathbf{z}_{c,i}, 
    \label{eq:CF_output}
\end{aligned}
\end{equation}
where $\odot$ is the element-wise multiplication, $\mathbf{F}_{i}$ is the same input as mentioned in eq~\eqref{eq:spatial_pool}, fed to the channel fusion module corresponding to each modality, and $\mathbf{F}^{out}_{c,i}\in \mathbb{R}^{NT\times C_{i}\times H\times W}$ has the same dimension as the input $\mathbf{F}_{i}\in \mathbb{R}^{NT\times C_{i}\times H\times W}$.

\noindent
\textbf{Spatial Fusion.}\hspace{3pt} The channel fusion ignores local spatial information, which will be addressed by the spatial fusion module as seen in Figure~\ref{fig:team_SF}. We first globally average the output $\mathbf{F}^{out}_{c,i}$ of the channel fusion module over the channel dimension
\begin{equation}
    \mathbf{F}_{s,i} = \frac{1}{C_i}\sum^{C_i}_{j=1}\mathbf{F}^{out}_{c,i}[:,j,:,:], 
    \label{eq:channel_pool}
\end{equation}
where $\mathbf{F}_{s,i} \in \mathbb{R}^{NT\times 1\times H\times W}$. We then concatenate these three feature maps over the channel dimension to get a joint spatial representation $\mathbf{z}_{s}\in \mathbb{R}^{NT\times 3\times H\times W}$. We generate three spatial attention maps for each modality using the joint representation $\mathbf{z}_{s}$ 
\begin{equation}
    \mathbf{z}_{s,i} = \sigma(\mathbf{W}_{s,i}*\mathbf{z}_{s}), 
    \label{eq:SF_attention_map}
\end{equation}
where $\mathbf{z}_{s,i}\in \mathbb{R}^{NT\times 1\times H\times W}$, $\mathbf{W}_{s,i}$ is a $3\times 3$ 2D convolution kernel for modality $i$, and $*$ is the convolution operation. Similar to the channel fusion, the final output of the spatial fusion can be computed as
\begin{equation}
    \mathbf{F}^{out}_{s,i} = \mathbf{F}^{out}_{c,i} \odot \mathbf{z}_{s,i}, 
    \label{eq:SF_output}
\end{equation}
where $\mathbf{F}^{out}_{s,i}\in \mathbb{R}^{NT\times C_{i}\times H\times W}$ has the same dimension as the input $\mathbf{F}_{i}\in \mathbb{R}^{NT\times C_{i}\times H\times W}$. 

\noindent
\textbf{Instantiations.}\hspace{3pt} The TEAM module is composed of the channel and the spatial fusion in a sequential manner as seen in Figure~\ref{fig:team_module}. TEAM ingests three modalities and outputs interacted features for each modality. Our idea of TEAM is generic, and it can be instantiated with different backbones (e.g.,~\citep{he2016deep,szegedy2017inception,sandler2018mobilenetv2}) and action recognition implementation specifics (e.g.,~\citep{wang2016temporal,lin2019tsm,li2020tea,Wang_2021_CVPR}). In this work, we demonstrate instantiation of our TEAM-Net on the ResNet architecture for action recognition. TEAM is inserted after $\text{Conv}_1$, $\text{Res}_2$, $\text{Res}_3$ and $\text{Res}_4$ respectively. Details of the instantiation can be referred to Figure~\ref{fig:TEAM_net} and Table~\ref{tab:sup_resnet_architecture} in \textit{Supplementary Materials}~\ref{sup-sec:backbone}.

\section{Experiments}
\subsection{Setups}
\noindent
\textbf{Datasets.}\hspace{3pt} We evaluate the performance of the proposed TEAM-Net on three public action recognition datasets Kinetics-400~\citep{carreira2017quo}, UCF-101~\citep{soomro2012ucf101} and HMDB-51~\citep{kuehne2011hmdb}. Kinetics-400 contains 400 human action categories and provides URL links for \textasciitilde 240k training videos and \textasciitilde 20k validation videos. We successfully collected 213,991 training videos and 17,575 validation videos since around 10\% of URLs are no longer valid. We directly report the accuracy on the validation set for Kinetics-400. UCF-101 contains 101 classes with 13,320 videos. HMDB-51 includes 51 classes with 6,766 videos. For these two datasets, we follow previous works~\citep{wu2018compressed,huo2020lightweight,li2020slow} to utilize three splits for training and evaluation. The average accuracy on three splits is reported. Following previous works~\citep{wu2018compressed,shou2019dmc}, we use MPEG-4 encoded videos, which have on average 11 P-frames for every I-frame in each GOP.

\noindent
\textbf{Training and Inference.}\hspace{3pt} We prepared partially decoded videos to conduct our experiments on video action recognition tasks by using the same encoding tool as CoViAR~\citep{wu2018compressed}. Given a partially decoded video input, we firstly divided it into $T$ segments of equal duration. Then we randomly selected one GOP from each segment to obtain a clip input with $T$ GOPs. The first frame in each GOP was chosen as an I-frame. The MV and residual were obtained from a randomly picked P-frame in each GOP. In other words, one partially decoded video clip input contains $T$ I-frames, $T$ MVs and $T$ residuals to be fed to the network. The size of the shorter side of these frames was fixed to 256, and corner cropping and scale-jittering were utilized for data augmentation. Each cropped frame was finally resized to $224\times 224$ for training the network. We utilized 2D ResNet-50 as a backbone for the I-frame modality, and 2D ResNet-18 as backbones for MV and residual modalities for both TEAM-Net and CoViAR baseline. For TSN, we used 2D ResNet-50 as the backbone. Models were trained on a NVIDIA DGX station with four Tesla V100 GPUs. We adopted SGD as the optimizer with a momentum of 0.9 and a weight decay of $5\times 10^{-4}$. Batch size was set to $N = 80$. For Kinetics, network weights were initialized using ImageNet pretrained weights. We started an initial learning rate of 0.01 and reduced it by a factor of 10 at 30, 40, 45 epochs and stopped at 50 epochs. For UCF-101 and HMDB-51, the network was finetuned using a pretrained Kinetics model. We started an initial learning rate of 0.005 and reduced it by a factor of 10 at 15, 20, 25 and stopped at 30 epochs. Following previous works~\citep{wu2018compressed}, we used 25 segments with a center-crop strategy for inference.

\subsection{Improving Baselines}
\noindent
\textbf{Action Recognition Performance.}\hspace{3pt} We compare TEAM-Net with two fundamental baselines with full decoding (TSN) and partial decoding (CoViAR) on three datasets in Table~\ref{tab:2D_baseline_compare}. It can be noticed that both TEAM-Net and CoViAR are able to improve the performance on UCF-101 and HMDB-51, which indicates that MVs and residuals contained in the partially decoded videos help the action reasoning. Interestingly, we observe there is a performance drop (0.9\%) of CoViAR on Kinetics compared to TSN. On the contrary, our TEAM-Net is still able to improve the performance by 2.2\% compared to TSN. This demonstrates that TEAM-Net is more robust on different datasets compared to CoViAR. Our findings here are consistent with previous research~\citep{wu2018compressed,wang2016temporal} that proves consecutive frames are highly redundant in videos. Proper use of P-frames in partially decoded videos is able to boost the model performance~\citep{wu2018compressed,huo2020lightweight,li2020slow,shou2019dmc}.

\noindent
\textbf{Computational Efficiency.}\hspace{3pt} Current frameworks~\citep{wang2016temporal,li2020tea,Wang_2021_CVPR} usually sample frames sparsely from videos, that means there is no significant difference on the computation when the model ingests raw videos or partially decoded videos i.e., the number of sampled frames is same if excluding P-frames. However, the computation is a bit heavier when taking P-frames into account. Table~\ref{tab:2D_baseline_compare} shows the inference speed of TEAM-Net and two baselines. TSN is about $1.5$ times faster compared to TEAM-Net. TEAM-Net performs closely to CoViAR but saves the training cost significantly \textit{i.e., only need to train one model instead of three}.
\begin{table}[!htbp]
    \centering
    \caption{TEAM-Net consistently outperforms baselines on three representative datasets. All methods use 8 frames for fair comparison. 
    }
    \setlength{\tabcolsep}{9mm}{
    \begin{threeparttable}
    \begin{tabular}{ccccc}
        \toprule
         Model& Kinetics-400& UCF-101 & HMDB-51 & Speed\\
         \midrule
         TSN\tnote{1} &  70.0& 89.7 & 66.8 & \textbf{45} V/s\\
        CoViAR\tnote{2} & 69.1 (\textcolor{blue}{\textbf{$\downarrow$}0.9})& 91.0 (\textcolor{red}{\textbf{$\uparrow$}1.3})& 73.2 (\textcolor{red}{\textbf{$\uparrow$}6.4}) & 31 V/s\\ 
        \cellcolor{columbiablue}TEAM-Net & \cellcolor{columbiablue}\textbf{72.2} (\textcolor{red}{\textbf{$\uparrow$}2.2})&\cellcolor{columbiablue}\textbf{94.3} (\textcolor{red}{\textbf{$\uparrow$}4.6})&\cellcolor{columbiablue}\textbf{73.8} (\textcolor{red}{\textbf{$\uparrow$}7.0}) & \cellcolor{columbiablue}30 V/s\\
         \bottomrule
    \end{tabular}
  \begin{tablenotes}
     \item[1] Different from~\citep{wang2016temporal}, TSN on UCF-101 and HMDB-51 are finetuned using a pretrained model on Kinetics-400 in this work, which give higher accuracies compared to~\citep{wang2016temporal}.
     \item[2] We re-implemented CoViAR using the official code in~\citep{wu2018compressed} by replacing ResNet-152 with ResNet-50 for the I-frame modality for fair comparison.
  \end{tablenotes}
    \end{threeparttable}
    \label{tab:2D_baseline_compare}
    }
\end{table}

\begin{table}[!htbp]
    \centering
    \caption{Comparison with state-of-the-arts using partially decoded videos on Kinetics, UCF and HMDB. Our TEAM-Net is an end-to-end (ETE) network and optical flow (OF) free.}
    \label{tab:SOTAs_compare}
    \setlength{\tabcolsep}{1mm}{
    \begin{threeparttable}
    \begin{tabular}{c|c|c|c|cccccc}
        \toprule
         \multicolumn{1}{c}{Method}& \multicolumn{1}{c}{OF}& \multicolumn{1}{c}{ETE}& \multicolumn{1}{c}{Backbone}& Kinetics& UCF& HMDB \\ 
         \midrule
         DMC-Net~\citep{shou2019dmc}& \cmark& \xmark&ResNet152 (I), ResNet18 (P)& -& 92.3&71.8 \\ \hline
         ATTP~\citep{huo2020lightweight} & \cmark& \xmark&EfficientNet (I), EfficientNet (P)& -& 91.1&62.9 \\ \hline 
         IMRNet~\citep{IMRNet}&  \cmark& \cmark& 3D-ResNet50 (I), 3D-ResNet50 (P)& -& \textbf{95.1}&72.2 \\ \hline \hline
         CoViAR~\citep{wu2018compressed}& \xmark& \xmark&ResNet50 (I), ResNet18 (P)& 69.1& 91.0&73.2 \\ \hline
         SIFP~\citep{li2020slow}& \xmark& \cmark&SlowFast-ResNet50& -& 94.0&72.3 \\ \hline
         MFCD-Net~\citep{battash2020mimic}&  \xmark& \cmark& Multi-Fiber Network~\citep{chen2018multi}& 68.3& 93.2&66.9 \\ \hline
         IMRNet~\citep{IMRNet}&  \xmark& \cmark& 3D-ResNet50 (I), 3D-ResNet50 (P)& -& 92.6&67.8 \\ \hline
        \rowcolor{columbiablue} TEAM-Net& \xmark& \cmark&ResNet50 (I), ResNet18 (P)& \textbf{72.2}& 94.3& \textbf{73.8}\\  
         \bottomrule
    \end{tabular}
    \end{threeparttable}
    }
\end{table}

\subsection{Comparison with State-of-the-Arts}
We compare our approach with current state-of-the-arts in the area of action recognition using partially decoded videos on Kinetics-400, UCF-101 and HMDB-51 in Table~\ref{tab:SOTAs_compare}. Unlike recent works~\citep{IMRNet,li2020slow,huo2020lightweight} that utilize advanced backbones such as 3D CNN~\citep{hara2018can}, SlowFast Networks~\citep{feichtenhofer2019slowfast} and EfficientNet~\citep{tan2019efficientnet}, it is worth nothing that our TEAM-Net utilizes 2D ResNet as backbones for three modalities. TEAM-Net learns the informatively fused features from I-frames and P-frames through the TEAM module. More advanced backbones are able to boost the performance for TEAM-Net. Our TEAM-Net, together with SIFP~\citep{li2020slow}, MFCD-Net~\citep{battash2020mimic} and IMRNet~\citep{IMRNet}, enjoys end-to-end training. It can be seen that our TEAM-Net outperforms current state-of-the-arts without optical flow and achieves competitive results even compared to those approaches with optical flow. Regarding HMDB-51, TEAM-Net outperforms all previous approaches with/without optical flow. In terms of UCF-101, TEAM-Net performs closely to IMRNet (3D CNN backbone) with optical flow. Only few methods are evaluated on Kinetics-400 since many approaches suffer from expensive training i.e., require training three separated backbones for each modality. Our TEAM-Net also performs well on Kinetcs-400, which demonstrates good ability for generalization across datasets.

\subsection{Ablation Study}
\noindent
\textbf{Arrangements of Channel and Spatial Fusion.}\hspace{3pt} We investigate different arrangements of channel and spatial fusion from a network engineering side. Table~\ref{tab:AB_arrangement} shows 5 possible arrangements for formulating a TEAM module. $\text{TEAM}_c$ and $\text{TEAM}_s$ indicate channel fusion only or spatial fusion only contained in the module. $\text{TEAM}_{c+s}$ refers to connecting channel and spatial module in a parallel way. $\text{TEAM}_{c\to s}$ indicates channel fusion followed by spatial fusion, vice versa for $\text{TEAM}_{s \to c}$. It can be seen that three possible combinations of channel and spatial fusion work better than channel fusion only and spatial fusion only on both UCF and HMDB in Table~\ref{tab:AB_arrangement}. The sequential arrangement, in which the channel fusion is followed by the spatial fusion $\text{TEAM}_{c\to s}$, performs the best compared to the other two.
\begin{table}[!htbp]
    \centering
    \caption{Arrangements of channel and spatial fusion. Performances are evaluated on UCF and HMDB using 8-frame inputs for training. Average accuracy for three splits is reported.}
    \begin{threeparttable}
    \begin{tabular}{cccccc}
        \toprule
         Dataset& $\text{TEAM}_c$ & $\text{TEAM}_s$ & $\text{TEAM}_{c+s}$ & $\text{TEAM}_{s \to c}$ & $\text{TEAM}_{c\to s}$\\
         \midrule
         UCF-101 & 93.7& 93.7& 94.0& 94.1& \textbf{94.3}\\ 
         HMDB-51 & 73.0& 72.9& 73.1& 73.2& \textbf{73.8}\\ 
         \bottomrule
    \end{tabular}
    \end{threeparttable}
    \label{tab:AB_arrangement}
    % }
\end{table}

\vspace{-15pt}
\begin{table}[!htbp]
    \centering
    \caption{Performance of different embedded locations.}
    \begin{threeparttable}
    \begin{tabular}{cccccc}
        \toprule
         Dataset & $\text{Stage}_{\{1\}}$& $\text{Stage}_{\{1,2\}}$ & $\text{Stage}_{\{1,2,3\}}$ & $\text{Stage}_{\{1,2,3,4\}}$ & $\text{Stage}_{\{1,2,3,4,5\}}$\\
         \midrule
         UCF-101 & 93.6& 93.2& 93.3& \textbf{94.3}& 91.7\\ 
         HMDB-51 & 72.8& 73.1& 72.7& \textbf{73.8}& 70.4\\ 
         \bottomrule
    \end{tabular}
    \end{threeparttable}
    \label{tab:AB_location}
    % }
\end{table}
\vspace{-5pt}
\noindent
\textbf{Location of TEAM.}\hspace{3pt} The default embedded locations of TEAM are after $\text{Conv}_1$, $\text{Res}_2$, $\text{Res}_3$ and $\text{Res}_4$ respectively i.e., $\text{Stage}_{\{1,2,3,4\}}$, which is shown in Figure~\ref{fig:TEAM_net} in \textit{Supplementary Materials}~\ref{sup-sec:backbone}. We find that including the last stage fusion ($\text{Res}_5$) deteriorates the fusion performance. We suppose this may be caused by the fact that the spatial resolution is too low at the last stage, which causes inaccurate spatial fusion and degrades the performance. The location $\text{Stage}_{\{1,2,3,4\}}$ works the best, which indicates the efficacy of the mid-level fusion for these three modalities. 

\begin{figure}[!htbp]
    \centering
    \subfigure[\tt{Cricket Shot}]{
        \includegraphics[width=.7\textwidth]{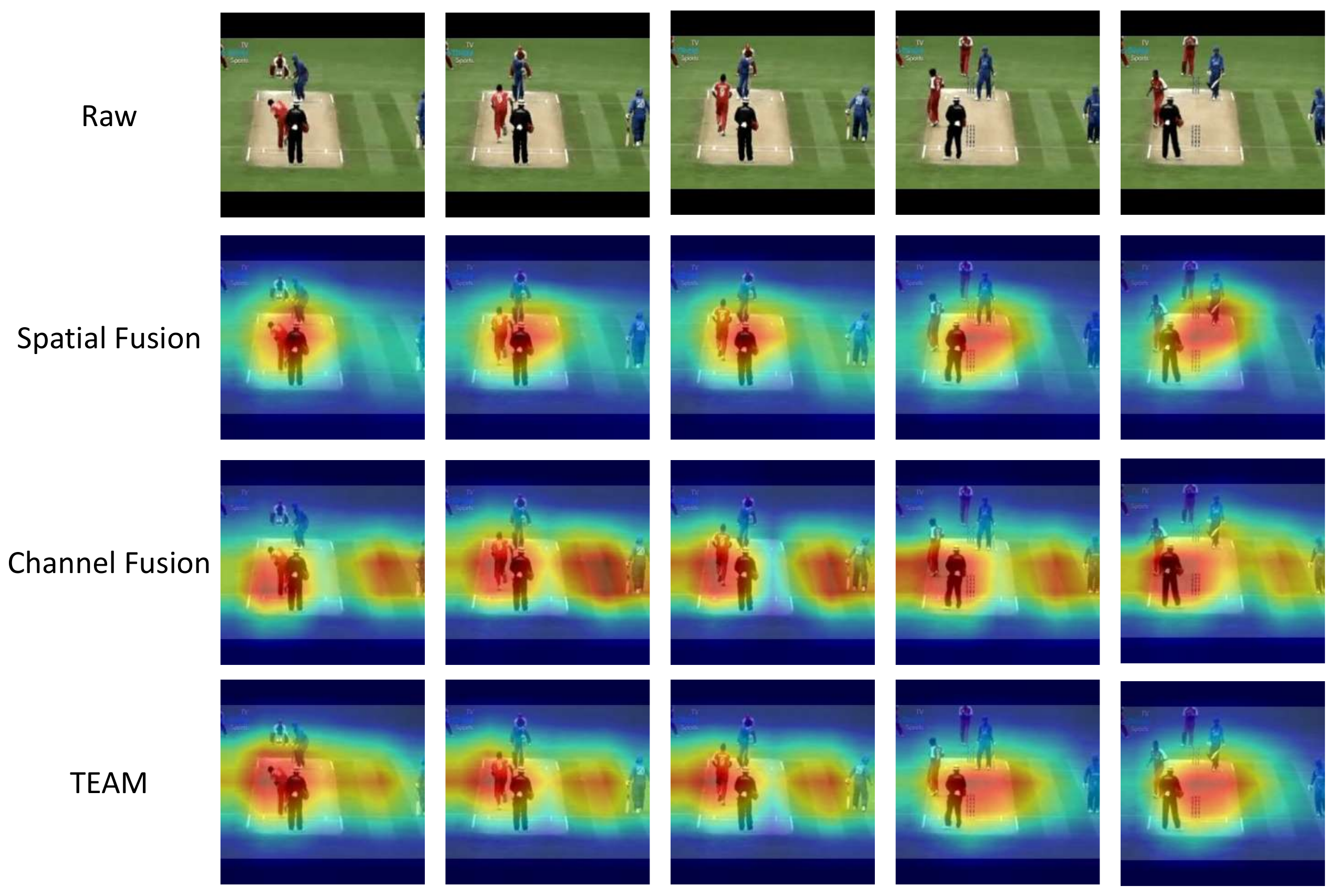}
        \label{fig:CricketShot_C}   
    }
    \subfigure[\tt{Pizza Tossing}]{
        \includegraphics[width=.7\textwidth]{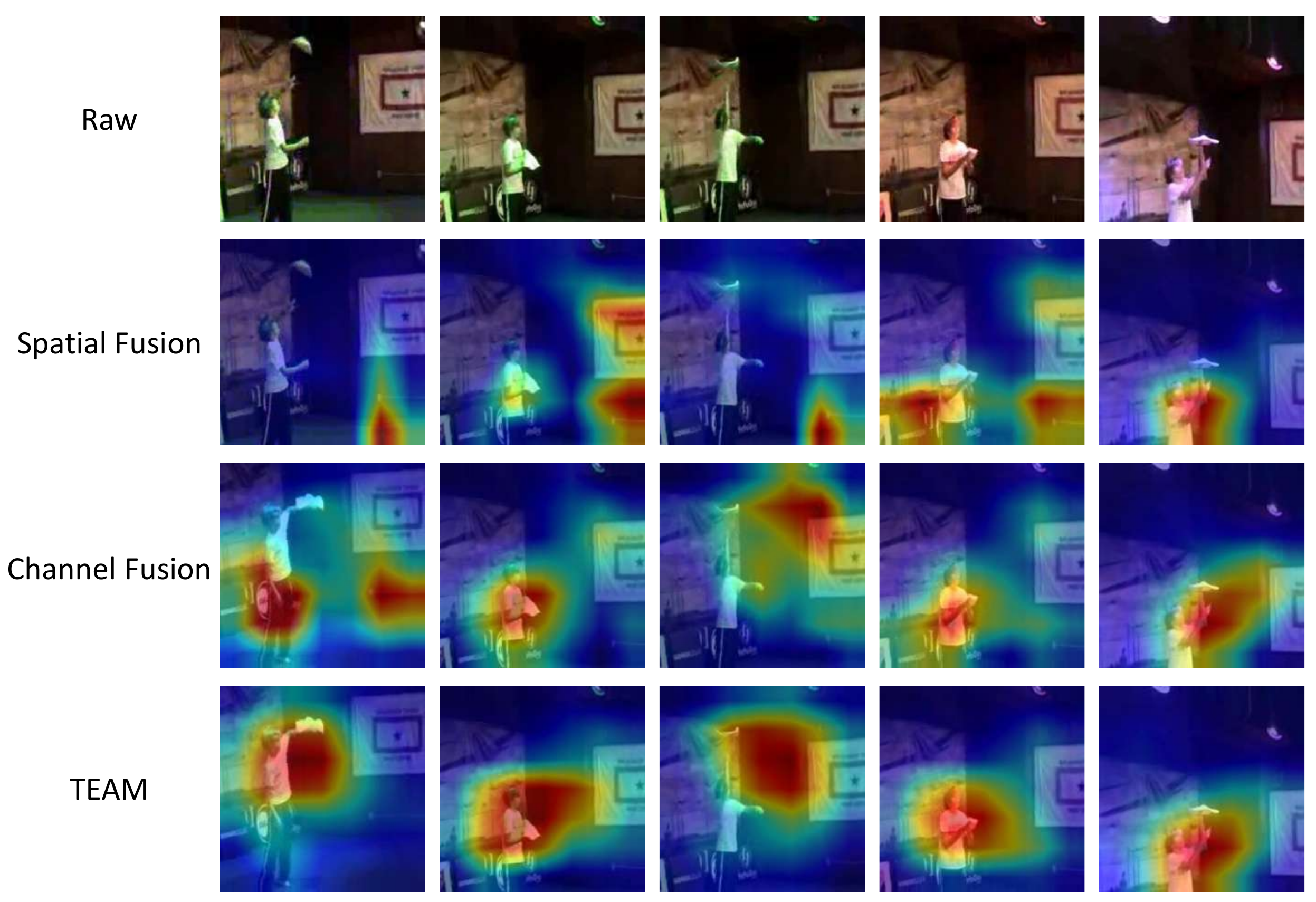}
        \label{fig:PizzaTossing_S}     
    }
    \caption{Class-specific visualization for channel fusion \textit{only}, spatial fusion \textit{only} and our proposed TEAM module. }
    \label{fig:S_C_vis}
\end{figure}

\noindent
\textbf{Understanding of TEAM.}\hspace{3pt} Previous experimental results show the promising performance of TEAM-Net. Here we provide insights behind the TEAM-Net and the effectiveness of the two fusion modules. Regarding the channel fusion, it takes global spatial information (see eq~\eqref{eq:spatial_pool}) into account to fuse each modality i.e., it focuses on \textit{`what'} is meaningful given a frame. This indicates that the channel fusion has low spatial resolution, which results in inaccurate local information reasoning. As seen on the 3rd row in Figure~\ref{fig:CricketShot_C}, there is a strong activation on the right for each frame detected by the channel fusion, while the main action is happening in the middle. The spatial fusion and TEAM characterize these actions happening in the local spatial space successfully, which suggests that spatial fusion focuses on \textit{`where'} is an informative part. However, it could be sensitive to unrelated movements in video. For instance, in the 2nd row in Figure~\ref{fig:PizzaTossing_S}, the spatial fusion falls into paying attention to the edge on several frames, which is probably caused by camera movement. Our proposed TEAM module takes \textit{`what'} and \textit{`where'} into account while fusing different modalities by incorporating two modules. Both recognition performance and visualization results confirm the effectiveness of the proposed approach. We provide more visualization results in \textit{Supplementary Materials}~\ref{sup-sec:vis}.    

\section{Conclusion}
In this work, we visit video action recognition in the partially compressed domain. We introduce a sampling strategy for inputs that comprises I-frames, MVs and residuals based on GOPs. We propose a novel TEAM module that is able to fuse three modalities effectively. The TEAM module comprises the channel fusion and the spatial fusion sequentially, and is embedded to the network in a plug-and-play manner as instantiations of TEAM-Net,  which enables training an end-to-end network for three modalities. We demonstrate that TEAM-Net outperforms the baseline TSN in the raw domain and current state-of-the-arts in the partially compressed domain on all three public datasets presented in this work. 

\section{Acknowledgment}
This work is financially supported by Science Foundation Ireland (SFI) under the Grant Number 15/RP/2776. We gratefully acknowledge the support of NVIDIA Corporation with the donation of the NVIDIA DGX station used for this research. The authors appreciate informative and insightful comments provided by anonymous reviewers, which significantly improve the quality of this work.

\clearpage
\bibliographystyle{unsrtnat}
\bibliography{references.bib}

\setcounter{figure}{0}
\setcounter{table}{0}
\setcounter{section}{0}
\clearpage
\setcounter{page}{1}
\renewcommand\thesection{\Alph{section}}
\section{Instantiations Architecture}\label{sup-sec:backbone}
\begin{figure}[!htbp]
    \centering
    \includegraphics[width=1.\textwidth]{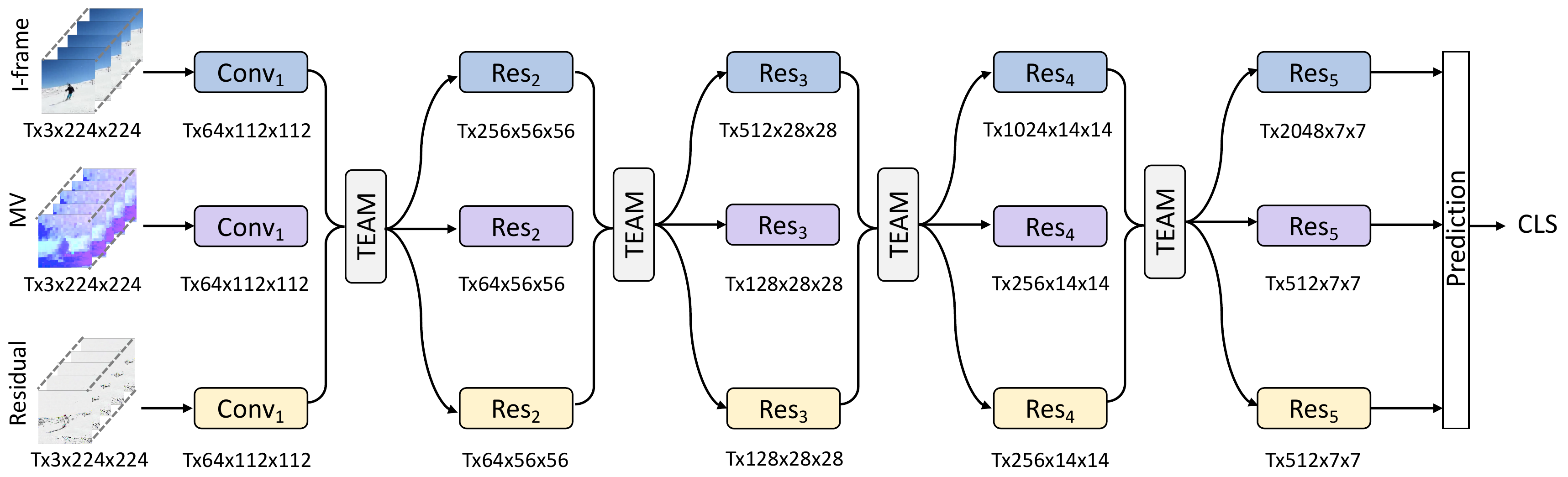}
    \caption{An example instantiation of the TEAM-Net. The dimensions of features at each stage are denoted as \{$T\times C \times H\times W$\}. The I-frame path ultilizes ResNet-50 while MV and residual paths use ResNet-18. }
    \label{fig:TEAM_net}
\end{figure}

In this work, our TEAM is instantiated with ResNet architecture to formulate TEAM-Net. I-frame pathway utilizes ResNet-50 as the backbone while MV and residual use ResNet-18 as the backbone. Details of the TEAM-Net architecture and feature dimension at each stage can be referred in Figure~\ref{fig:TEAM_net} and Table~\ref{tab:sup_resnet_architecture}. As mentioned in the main paper, the input is sampled based on the GOP. In Figure~\ref{fig:TEAM_net}, $T$ GOPs are sampled from a video, which results in $T$ I-frames and $T$ P-frames (MVs and residuals) to be fed to the network.  

\section{Instantiated with TSM}\label{sup-sec:TSM}
Table~\ref{tab:sup_TSM} shows the TEAM-Net using TSM as the backbone. It can be seen that TEAM is able to improve the TSM baseline consistently. Together with TSN experiment demonstrated in the main paper, these results show that TEAM-Net is well-generalized to different backbones and datasets.

\begin{table}[!htbp]
    \centering
    \caption{TEAM-Net consistently improves the TSM baseline for three datasets. All methods use 8 frames for fair comparison. 
    }
    \setlength{\tabcolsep}{7mm}{
    \begin{threeparttable}
    \begin{tabular}{cccc}
        \toprule
         Model& Kinetics-400& UCF-101 & HMDB-51\\
         \midrule
         TSM &  71.8& 94.5& 70.7 \\
        \cellcolor{columbiablue}TEAM-Net & \cellcolor{columbiablue}\textbf{73.0} (\textcolor{red}{\textbf{$\uparrow$}1.2})&\cellcolor{columbiablue}\textbf{95.2} (\textcolor{red}{\textbf{$\uparrow$}0.7})&\cellcolor{columbiablue}\textbf{74.3} (\textcolor{red}{\textbf{$\uparrow$}3.0}) \\
         \bottomrule
    \end{tabular}
    \end{threeparttable}
    \label{tab:sup_TSM}
    }
\end{table}

\begin{table}[!htpb]
    \centering
    \caption{TEAM-Net is instantiated with the ResNet architecture. I-frame pathway utilizes ResNet-50 as the backbone while MV and residual (R) both utilize ResNet-18 as the backbone.}
    \begin{tabular}{c|c|c|c}
        \toprule
        Stage & I-frame & MV/Residual & Output size \\ \hline
        \multirow{2}{*}{Input} & \multicolumn{2}{c|}{\multirow{2}{*}{——}} & I-frame: $T\times 224 \times 224$ \\ 
        & \multicolumn{2}{c|}{}& MV/R: $T\times 224 \times 224$\\
        \hline
        \multirow{2}{*}{$\text{conv}_1$} & \multicolumn{2}{c|}{\multirow{2}{*}{$1\times7\times 7$, 64, stride 1, 2, 2}} & I-frame: $T\times 112 \times 112$ \\ 
        & \multicolumn{2}{c|}{}& MV/R: $T\times 112 \times 112$\\
        \hline
        \multirow{2}{*}{$\text{pool}_1$} & \multicolumn{2}{c|}{\multirow{2}{*}{$1\times3\times 3$, max, stride 1, 2, 2}} & I-frame: $T\times 56 \times 56$ \\ 
        & \multicolumn{2}{c|}{}& MV/R: $T\times 56 \times 56$\\
        \hline
        \multicolumn{3}{c|}{\multirow{2}{*}{TEAM}} & I-frame: $T\times 56 \times 56$\\ 
        \multicolumn{3}{c|}{}& MV/R: $T\times 56 \times 56$\\ \hline
        $\text{res}_2$& $\Bigg[\begin{array}{c}
                                1\times1\times1, 64\\
                                1\times3\times3, 64\\
                                1\times1\times1, 256\\
                              \end{array}  
                        \Bigg]\times3$ &
                        $\Bigg[\begin{array}{c}
                                1\times3\times3, 64\\
                                1\times3\times3, 64\\
                              \end{array}  
                        \Bigg]\times2$ &
                         \makecell[c]{ \vtop{\hbox{\strut I-frame: $T\times 56 \times 56$}
                                              \hbox{\strut MV/R: $T\times 56 \times 56$}} }  \\ \hline
        \multicolumn{3}{c|}{\multirow{2}{*}{TEAM}} & I-frame: $T\times 56 \times 56$\\ 
        \multicolumn{3}{c|}{}& MV/R: $T\times 56 \times 56$\\ \hline
        $\text{res}_3$& $\Bigg[\begin{array}{c}
                                1\times1\times1, 128\\
                                1\times3\times3, 128\\
                                1\times1\times1, 512\\
                              \end{array}  
                        \Bigg]\times4$ &
                        $\Bigg[\begin{array}{c}
                                1\times3\times3, 128\\
                                1\times3\times3, 128\\
                              \end{array}  
                        \Bigg]\times2$ &
                        \makecell[c]{ \vtop{\hbox{\strut I-frame: $T\times 28 \times 28$}
                                              \hbox{\strut MV/R: $T\times 28 \times 28$}} } \\ \hline
        \multicolumn{3}{c|}{\multirow{2}{*}{TEAM}} & I-frame: $T\times 28 \times 28$\\ 
        \multicolumn{3}{c|}{}& MV/R: $T\times 28 \times 28$\\ \hline
        $\text{res}_4$& $\Bigg[\begin{array}{c}
                                1\times1\times1, 256\\
                                1\times3\times3, 256\\
                                1\times1\times1, 512\\
                              \end{array}  
                        \Bigg]\times6$ &
                        $\Bigg[\begin{array}{c}
                                1\times3\times3, 256\\
                                1\times3\times3, 256\\
                              \end{array}  
                        \Bigg]\times2$ &
                        \makecell[c]{ \vtop{\hbox{\strut I-frame: $T\times 14 \times 14$}
                                             \hbox{\strut MV/R: $T\times 14 \times 14$}} } \\ \hline
        \multicolumn{3}{c|}{\multirow{2}{*}{TEAM}} & I-frame: $T\times 14 \times 14$\\ 
        \multicolumn{3}{c|}{}& MV/R: $T\times 28 \times 28$\\ \hline
        $\text{res}_5$& $\Bigg[\begin{array}{c}
                                1\times1\times1, 512\\
                                1\times3\times3, 512\\
                                1\times1\times1, 2048\\
                              \end{array}  
                        \Bigg]\times3$ &
                        $\Bigg[\begin{array}{c}
                                1\times3\times3, 512\\
                                1\times3\times3, 512\\
                              \end{array}  
                        \Bigg]\times2$ &
                        \makecell[c]{ \vtop{\hbox{\strut I-frame: $T\times 7 \times 7$}
                                             \hbox{\strut MV/R: $T\times 7 \times 7$}} } \\ \hline
       \multicolumn{3}{c|}{global average pool, FC}& $T\times CLS$\\ \hline
       \multicolumn{3}{c|}{temporal average}& $CLS$\\
       \bottomrule
    \end{tabular}
    \label{tab:sup_resnet_architecture}
\end{table}

\newpage
\section{Class-specific Visualization}\label{sup-sec:vis}
\begin{figure}[!htpb]
    \centering
    \subfigure[\tt{Cricket Shot}]{
        \includegraphics[width=.7\textwidth]{images/CricketShot_C_compressed.pdf}
        \label{fig:sup_CricketShot_C}   
    }
    \subfigure[\tt{Hair Cut}]{
        \includegraphics[width=.7\textwidth]{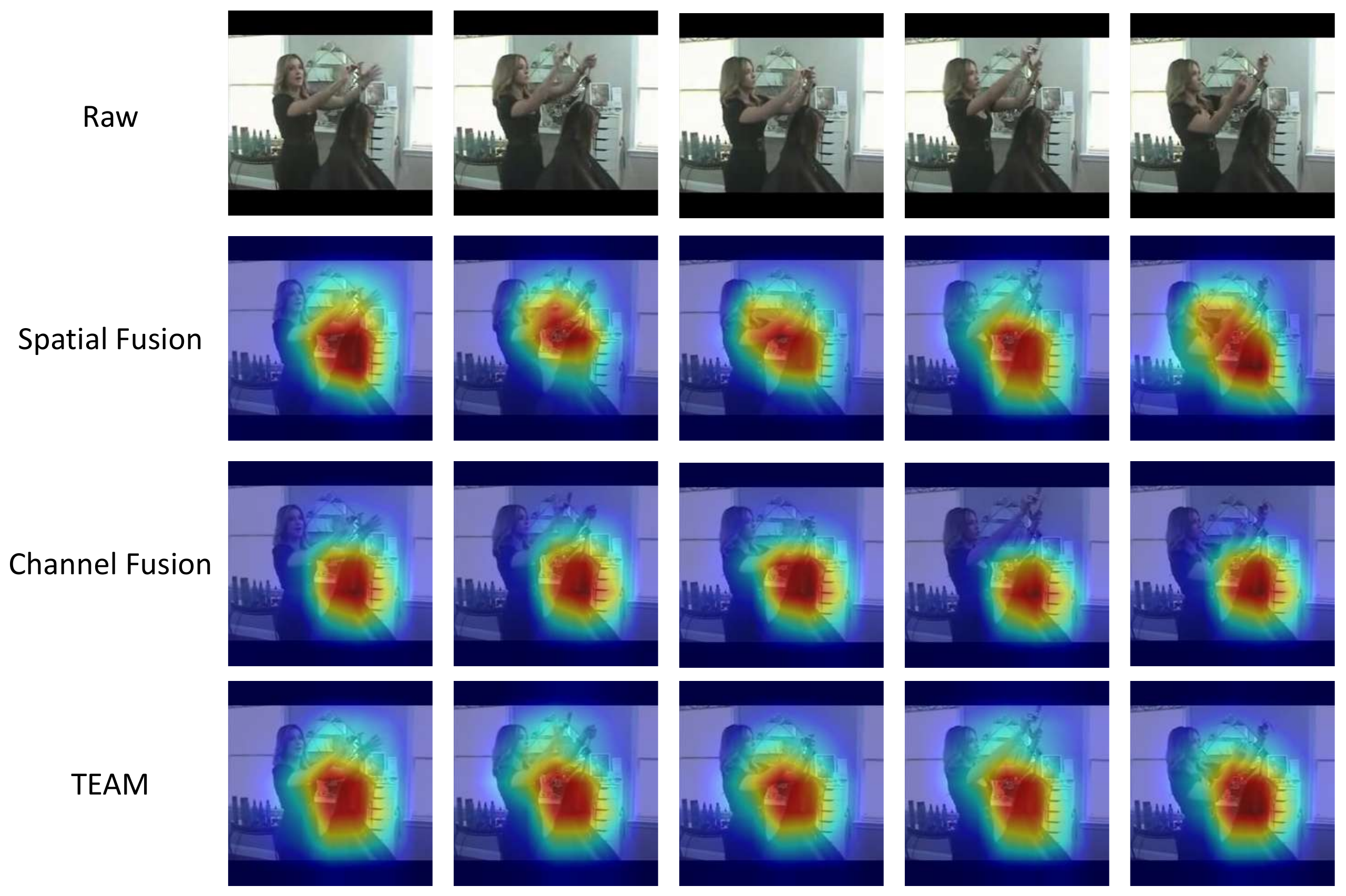}
        \label{fig:sup_Haircut_S}     
    }
    \caption{Class-specific visualization for channel fusion \textit{only}, spatial fusion \textit{only} and our proposed TEAM module. \textit{The channel fusion only reasoning is inaccurate in this case.}}
    \label{fig:sup_C_vis}
\end{figure}

\begin{figure}[!htpb]
    \centering
    \subfigure[\tt{Pizza Tossing}]{
        \includegraphics[width=.7\textwidth]{images/PizzaTossing_S_compressed.pdf}
        \label{fig:sup_PizzaTossing_S}     
    }
    \subfigure[\tt{Sky Diving}]{
        \includegraphics[width=.7\textwidth]{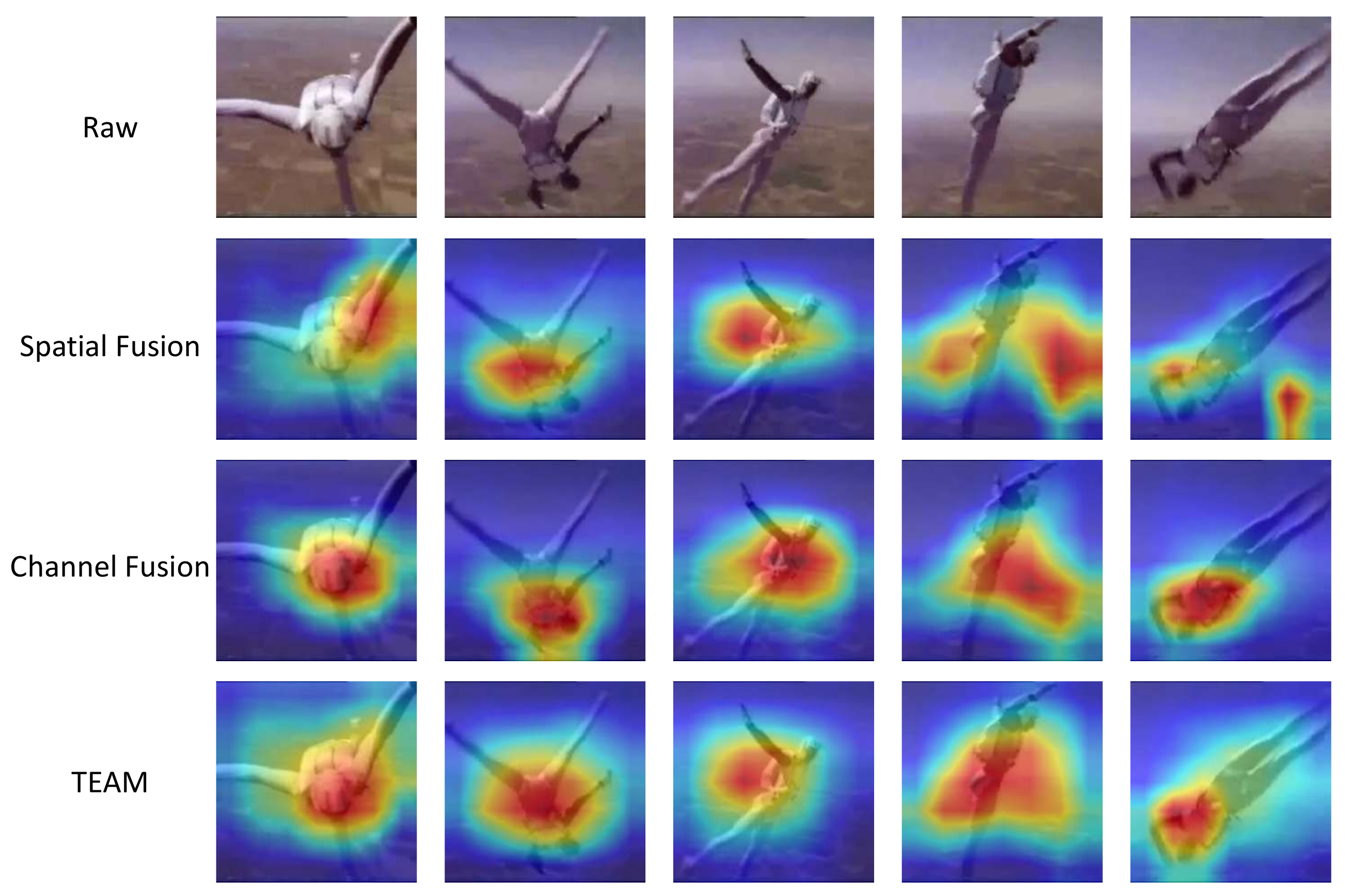}
        \label{fig:sup_SkyDiving_S}     
    }
    \caption{Class-specific visualization for channel fusion \textit{only}, spatial fusion \textit{only} and our proposed TEAM module. \textit{The spatial fusion only is sensitive to unrelated movements.}}
    \label{fig:sup_S_vis}
\end{figure}

As mentioned in the main paper, the channel fusion focuses on \textit{`what'} by taking global spatial information into account while the spatial fusion focuses on \textit{`where'} by looking at the local spatial content. Figure~\ref{fig:sup_C_vis} shows that the channel fusion fails in local action reasoning. Figure~\ref{fig:sup_S_vis} shows that the spatial fusion is sensitive to unrelated local movements.

\end{document}